\documentclass[pdflatex,sn-basic]{sn-jnl}
\let\oldbibliography\thebibliography
\renewcommand{\thebibliography}[1]{\oldbibliography{#1}
\setlength{\itemsep}{0pt}}
\newenvironment{where}{\noindent{}where,\begin{itemize}}{\end{itemize}}
\usepackage{graphicx}%
\usepackage{multirow}%
\usepackage{amsmath,amssymb,amsfonts}%
\usepackage{amsthm}%
\usepackage{mathrsfs}%
\usepackage[title]{appendix}%
\usepackage{textcomp}%
\usepackage{manyfoot}%
\usepackage{booktabs,tabularx}%
\usepackage{algorithm}%
\usepackage{algorithmicx}%
\usepackage{algpseudocode}
\usepackage{listings}%
\usepackage{caption}
\usepackage{subcaption}
\usepackage{threeparttable}
\usepackage{makecell}
\usepackage{cellspace} 
\usepackage{enumitem}
\usepackage{listings}
\usepackage{colortbl,xcolor}
\usepackage{fnpct}

\theoremstyle{thmstyleone}%
\theoremstyle{thmstyletwo}%

\theoremstyle{thmstylethree}%

\begin{document}

\title{Is it Still Fair? A Comparative Evaluation of Fairness Algorithms through the Lens of Covariate Drift}


\author[1,2,*]{\fnm{Oscar Blessed} \sur{Deho}}\email{oscar.deho@unisa.edu.au}

\author[2]{\fnm{Michael} \sur{Bewong}}\email{mbewong@csu.edu.au}

\author[2]{\fnm{Selasi} \sur{Kwashie}}\email{skwashie@csu.edu.au}
\author[1]{\fnm{Jiuyong} \sur{Li}}\email{jiuyong.li@unisa.edu.au}
\author[1]{\fnm{Jixue} \sur{Liu}}\email{jixue.liu@unisa.edu.au}
\author[1]{\fnm{Lin} \sur{Liu}}\email{lin.liu@unisa.edu.au}
\author[1]{\fnm{Srecko} \sur{Joksimovic}}\email{srecko.joksimovic@unisa.edu.au}
\affil[1]{\orgdiv{UnisA STEM}, \orgname{University of South Australia}, \orgaddress{\city{Adelaide}, \postcode{SA 5000}, \state{SA}, \country{Australia}}}
\affil[2]{\orgdiv{School of Computing, Mathematics and Engineering}, \orgname{Charles Sturt University}, \orgaddress{\city{Wagga Wagga}, \postcode{NSW 2650}, \state{NSW}, \country{Australia}}}
\affil[*]{Corresponding author: Oscar Blessed Deho oscar.deho@unisa.edu.au}


\abstract{Over the last few decades, machine learning (ML) applications have grown exponentially, yielding several benefits to society. However, these benefits are tempered with concerns of discriminatory behaviours exhibited by ML models. In this regard, fairness in machine learning has emerged as a priority research area. Consequently, several fairness metrics and algorithms have been developed to mitigate against discriminatory behaviours that ML models may possess. Yet still, very little attention has been paid to the problem of naturally occurring changes in data patterns (\textit{aka} data distributional drift), and its impact on fairness algorithms and metrics. In this work, we study this problem comprehensively by  analyzing 4 fairness-unaware baseline algorithms and 7 fairness-aware algorithms, carefully curated to cover the breadth of its typology, across 5 datasets including public and proprietary data, and evaluated them using 3 predictive performance and 10 fairness metrics.
In doing so, we show that (1) data distributional drift is not a trivial occurrence, and in several cases can lead to serious deterioration of fairness in so-called fair models; (2) contrary to some existing literature, the size and direction of data distributional drift is not correlated to the resulting size and direction of unfairness; and (3) choice of, and training of fairness algorithms is impacted by the effect of data distributional drift which is largely ignored in the literature. Emanating from our findings, we  synthesize several policy implications of data distributional drift on fairness algorithms that can be very relevant to stakeholders and practitioners.}




\maketitle

\section{Introduction}
Nowadays, machine learning (ML) models are ubiquitous in all human settings. They have even been used in judicial systems to determine whether or not to parole convicted criminals \citep{angwin16}. While ML systems have undoubtedly improved human productivity \citep{furman2019ai}, they can also propagate biases in data, consequently discriminating against certain demographic groups. Notably, COMPAS, a software for parole sentencing, was found to label black defendants twice as likely as their white counterparts to recidivate over a two-year period \citep{angwin16}. 
\par In recent years, there has been significant research on the so-called \textit{fairness in machine learning} which is primarily concerned with measuring and mitigating algorithmic biases\footnote{Throughout this work, we interchangeably use algorithmic bias, discrimination, or unfairness to refer to differential decisions made by an algorithm based on an individual's demographic attribute due to (or independent of) biases in the dataset.}. Several fairness algorithms constrained to satisfy certain fairness metrics have been designed \citep{mehrabi19}. Majority of existing fairness algorithms are designed on the assumption that the training and test datasets are \textit{independently and identically distributed (iid)} \citep{disgraph,maity2021does}. In other words, it is assumed that ensuring fairness on the training dataset approximates fairness guarantees on the test dataset. 
\par However in reality, it is common for the distribution of the test dataset (target environment) to drift from that of the training dataset (source environment), thus no longer \textit{identically} distributed. For example, a recent study demonstrated distributional drift of online learning behaviors within the educational setting at the height of COVID-19 \citep{impey2021moocs}. In such circumstances, we posit that fairness algorithms cannot guarantee fairness. In recent years, only a handful of research works investigate the robustness of fairness algorithms amidst distributional drifts \citep{kamp2021robustness}. Nonetheless, there remain gaps in the  literature. Firstly, the existing literature does not quantify the size of \textit{differential drift}\footnote{We refer to differential distributional drift as the \textit{disparity} in the \textit{magnitude} of the drift in data distribution across \textit{different} demographic groups.} in data distribution across demographic groups. Therefore, it is difficult to characterize the relationship between differential distributional drift and the corresponding fairness.  This characterization is needed in order to develop fairness algorithms that are robust against distributional drift. Secondly, there is no comprehensive study that benchmarks fairness algorithms vis-a-vis fairness metrics in the context of data distributional drift. Our study fills this gap by considering a large collection of metrics, algorithms and datasets and explores the conditions under which fairness algorithms may be considered fair amidst distributional drift. More formally, we explore the following research questions (RQs).
\begin{itemize}
     \item \textbf{RQ1}: What is the relationship between differential drift in data across demographic groups and algorithmic fairness?
    \item \textbf{RQ2}: Are existing fairness algorithms distributional drift aware?
\end{itemize}
By investigating these RQs, we can diagnose and reveal various ways that distributional drift impacts algorithmic fairness. To answer these questions, we perform an extensive experimental analysis involving 4 baseline models and 7 fairness algorithms using 3 predictive performance and 10 fairness metrics. We run our analysis on 5 real-world datasets across 4 domains: Education, Finance, Employment, and Criminal Justice. The \textbf{main contributions} of our work are as follows:
\begin{itemize}
    \item We reveal several interesting relationships between distributional drift---specifically covariate drift---and algorithmic fairness. Especially, we show how covariate drift results in discrimination (or reverse discrimination). 
    \item  We demonstrate the lack of robustness of existing algorithms in the face of covariate drift, while highlighting the need for the proper contextualization of what is \textit{fair}.  
    \item  We perform extensive experimental analysis that contributes critical empirical evidence on the impact of covariate drift on algorithmic fairness and recommend important policy implications of our findings for relevant stakeholders. 
\end{itemize}
The rest of this paper is organized as follows. Section \ref{sec:related works} discusses relevant related works. Section \ref{sec:prelims} provides preliminary information regarding notations, evaluation metrics, and fairness algorithms. Section \ref{sec:experiment} discusses our experimental setup. We discuss the results and their implications in Section \ref{sec:resdisc}. We conclude the paper in Section \ref{sec:conc}.

\section{Related Work}
\label{sec:related works}
Fairness algorithms either modify biased data (\textit{aka} pre-processing algorithms), add a fairness constraint directly to a model's objective function (\textit{aka} in-processing algorithms), or modify a biased model's outcomes (\textit{aka} post-processing algorithms). In recent years, a number of studies have been done to assess the various fairness algorithms based on different evaluation criteria \citep{freid,roth2018comparison,oscar}. Existing works tend to compare fairness algorithms in terms of their predictive performance, fairness, and/or fairness-accuracy tradeoff. For instance, \cite{hamilton2017benchmarking} compared four fairness algorithms using four fairness metrics across three different datasets. However, the author found that the fairness of the algorithms varied across datasets. In a similar study, \cite{roth2018comparison} compared three fairness algorithms and found that in most cases, the in-processing approaches tend to achieve fairness better than the pre-processing approaches. Nonetheless, Roth found that the algorithms were inconsistent across datasets and remarked the need for further extensive experiments. Furthermore, studies by \cite{freid} and \cite{oscar} have compared fairness algorithms across several datasets, investigated the effect of hyperparameter variation on fairness algorithms, and evaluated the fairness-accuracy trade-offs of various fairness algorithms. Both studies also found that the performance of fairness algorithms tends to vary across datasets. In this regard, a new line of research that is focused on a comparative analysis of the utility of fairness algorithms due to drifts in data distribution has recently emerged.
\par Only a handful of  work investigate the robustness of fairness algorithms in the face of distributional drift. For example, \cite{castelnovo2021towards} trained three fairness algorithms on banking data and observed that the fairness algorithms failed to satisfy demographic parity one year after deployment when the financial data drifted for certain demographic groups. Similarly, \cite{ghodsi2022context} trained one pre-processing and one in-processing algorithm on the newly released Adult Dataset and found that the fair models significantly deteriorate in terms of predictive performance and fairness due to drift in the spatial distribution of races. In what is perhaps the closest to our study, \cite{islam2022through} performed a comprehensive analysis of different fairness algorithms across different datasets and investigated the robustness of the algorithms to what they refer to as \textit{data errors}, which in some sense, amounts to distributional drift. Specifically, \cite{islam2022through} introduced data errors by (1) swapping certain column values, (2) scaling certain column values, and (3) imputing missing values for certain columns. All the data errors were introduced randomly and disproportionately across demographic groups. Interestingly, \cite{islam2022through} found that pre- and in-processing algorithms tend to be less generalizable, whereas post-processing algorithms were found to be more robust to data errors. In a recent related work, \cite{gardner2023cross}, performed cross-institutional analysis where learning analytics models from one institution were deployed in another institution. Contrary to prior findings, \cite{gardner2023cross} found almost no drop in the fairness of their models when deployed in a different institution where there is a likelihood of distributional drift. It is worth noting that the models used in the study by \cite{gardner2023cross}, however, were traditional ML models without any fairness constraints.
\par None of the related work measure the level of distributional drift in the data thus making it difficult to draw a relationship between distributional drift and fairness. Further, existing works that investigate the impact of distributional drift on fairness \citep{castelnovo2021towards,ghodsi2022context,kamp2021robustness} cover a very limited breadth in terms of fairness metrics and algorithms. In particular, they are often limited to three fairness metrics namely, equalized odds, statistical parity ({\textit{aka} demographic parity), and equal opportunity. However, given the fluidity of fairness metrics, accompanied by the impossibility theorem \citep{Chould}, it has become critical to have a more comprehensive benchmarking of fairness algorithms and notions through the lens of data distributional drift.

\section{Preliminaries and Background}
\label{sec:prelims}
Consistent with literature, we term attributes which are often not legally allowed to be used as the basis for decisions as protected attributes denoted $A$ e.g., race and gender \citep{siegel2003equality}. We represent non-protected attributes by $X$, actual outcomes by $Y$ and predicted outcomes by $\hat{Y}$.  We denote the privileged group and favourable outcome by $A=1$ and $Y=1$ (or $\hat{Y}=1$) respectively. Conversely, we represent unprivileged and unfavourable outcome by $A=0$ and $Y=0$ (or $\hat{Y}=0$) respectively. As per convention in algorithmic fairness literature, we denote demographic groups that are historically ``disadvantaged'' e.g., racial minorities as unprivileged groups. These groups are protected by law or simply referred to as protected groups \citep{mehrabi19}. Non-protected/privileged groups are the demographic groups that are historically advantaged e.g., racial majorities.

\subsection{ Types of Distributional Drift and Drift Detection Metric}
Given a non-protected input attribute $X$, and target outcome $Y$, at time $t$ (e.g., training), and time $t+k$ (e.g., testing), where $k >0$, distributional drift can manifest in the following 3 ways \citep{datashift}:
\begin{enumerate}[label=\alph*)]
  \item Covariate drift: This occurs when the \textit{input} marginal probability $P(x)$ changes but the conditional probability $P(y|x)$ does not change. Covariate drift is expressed as: 
 \begin{equation}
\footnotesize
P_{t}(x)\neq P_{t+k}(x) \land P_{t}(y|x)= P_{t+k}(y|x)
\end{equation} 
  \item Target drift: This occurs when the \textit{target} marginal probability $P(y)$ changes but the conditional probability $P(x|y)$ does not change. Target drift is expressed as:
\begin{equation}
\footnotesize
P_{t}(y)\neq P_{t+k}(y) \land P_{t}(x|y)= P_{t+k}(x|y)
\end{equation} 
  \item Concept drift: This occurs when the underlying relationship between the input and the target label changes but the input marginal probability $P(x)$ does not change. Concept drift is expressed as:
\begin{equation}
\footnotesize
P_{t}(y|x)\neq P_{t+k}(y|x) \land P_{t}(x)= P_{t+k}(x)
\end{equation} 
\end{enumerate}
Consistent with well-known related works \citep{maity2021does,yangXu}, we focus on the most prevalent drift, covariate drift. Further, we also make the assumption that the condition $P_{t}(y|x)= P_{t+k}(y|x)$ in covariate drift is satisfied. Consistent with \citep{yangXu}, this assumption is reasonable due to data insufficiency.
\par To measure covariate drift (i.e., $P_t(x) \neq P_{t+k}(x)$) condition, we adopt the well-known statistical drift detection metric called Jensen-Shannon Distance (JSD) \citep{jsd}. JSD works well for both categorical and numerical variables and is based on the KL divergence. However, unlike KL divergence, JSD is symmetric and returns  a finite score between 0 and 1. A drift value 0 signifies no drift and a drift value of 1 signifies maximum drift. More formally, given $P_t $ and $P_{t+k}$:

\begin{equation}
\label{eq:jsd}
\footnotesize
JSD(P_t,P_{t+k})= \sqrt{\frac{D_{KL}(P_t||M)+ D_{KL}(P_{t+k}||M)}{2}}
\end{equation}

\begin{where}
\item $M = \frac{P_t + P_{t+k}}{2}$
\item We consider $JSD \geq 0.1$ in this study  as a significant drift\footnote{There is no universally accepted threshold for significant drift. However, we chose 0.1 based on an extensive study by researchers at Evidently AI which is a drift monitoring company (\href{https://www.evidentlyai.com/blog/data-drift-detection-large-datasets}{https://www.evidentlyai.com/blog/data-drift-detection-large-datasets})}.
\end{where}
\subsection{Evaluation Metrics}
In this section we present the various predictive performance and fairness metrics considered in this work.
\subsubsection{Predictive Performance Metrics}
To measure the predictive performance of our models, we consider 3 popularly used metrics namely accuracy  for balanced data, and balanced accuracy and weighted F1-score for imbalanced data. All 3 metrics return a value of 0 (worst) to 1 (best).

\subsubsection{Fairness Metrics}
To measure the fairness of our models, we carefully consider 10 fairness metrics such that each of them belongs to at least one of the 7 clusters discovered from 26 fairness metrics by \cite{majumder2021fair}. The 7 clusters of metrics represent misclassification metrics (cluster 0,3), differential fairness metrics (cluster 1), individual fairness metrics (cluster 2), confusion matrix based group fairness metrics (cluster 4), between group individual fairness metrics (cluster 5) and intermediate metrics (cluster 6). Metrics in the same cluster were found to satisfy similar notions by the authors. Table \ref{tab:fairness_metrics} summarizes these metrics into two types described as follows:\\
\textit{Group Fairness Metrics.} Group fairness seeks to ensure that individuals belonging to different demographic groups are treated equally. The group fairness metrics that we considered in this work are statistical parity (SP) \citep{dwork2012fairness}, disparate impact (DI) \citep{feldman2015certifying}, error rate difference (ERD) \citep{berk2021fairness}, equalized odds (EO) \citep{hardt16}, equal opportunity(EOP) \citep{hardt16}, positive predictive value difference (PPV-DIFF) \citep{Chould}, and negative predictive value difference (NPV-DIFF) \citep{berk2021fairness}.\\
\textit{Individual Fairness Metrics.} Individual fairness is based on the idea that similar individuals should be treated similarly. There are fewer individual fairness metrics compared to group fairness metrics. The individual fairness metrics considered in this study are within group generalized entropy index (WGEI) \citep{spei}, within group theil index (WGTI) \citep{spei}, and \textit{consistency} \citep{zemel2013learning}. 

\begin{table*}[!htbp]

\caption{Formulae of various fairness metrics. }
\label{tab:fairness_metrics}
\centering
\tiny
 \begin{threeparttable}
\begin{tabularx}{\textwidth}{l|l|l|l|l}
\toprule
Metric & Formula & Type & Cluster& Range (Ideal Value) \\
\midrule
 DI        & $(P(\hat{Y}=1|A=0)\slash{P(\hat{Y}=1|A=1)}$  & G & 4,6 & [0, $\infty$] (1)   \\
 SP  & $P(\hat{Y}=1|A=0) - P(\hat{Y}=1|A=1)$& G & 4,1,6 &  [0,1](0)   \\

EO	&		$P(\hat{Y}=1|A=0, Y=y)-P(\hat{Y}= 1|A=1, Y=y), y \in \{0,1\}$		 & G&	4 & [0,1] (0)	\\

EOP	&				$P(\hat{Y}=1|A=0, Y=1)- P(\hat{Y}= 1|A=1, Y=1)$	  & G & 4 &[0,1] (0) \\

ERD &			$( (FP+FN)\slash (P+N))_{A=0}-((FP+FN)\slash(P+N))_{A=1}$	 & G & 0 & [0,1] (0) 	\\
PPV-DIFF 	&			$(TP\slash (TP+FP))_{A=0}-(TP\slash(TP+FP))_{A=1}$	 & G & 3 & [0,1] (0)	\\

NPV-DIFF&		$(TN\slash (TN+FN))_{A=0}-(TN\slash(TN+FN))_{A=1}$	 & G& 0 & [0,1] (0)\\	

WGEI	&		$\varepsilon^{\alpha}_{\omega}(\textbf{b})$, $\alpha\not\in\{0,1\}$	 & I & 5,2 & [0,1] (0) 	\\
WGTI &		$\varepsilon^{\alpha}_{\omega}(\textbf{b})$, 	$\alpha = 1$	 & I & 5,2 &  [0,1] (0) \\
Consistency & $1-\frac{1}{Nk} \sum_{i=1}^{N} \sum_{j \in k N N(X_i)}\left|\hat{Y}_i-\hat{Y}_j\right|$  & I  & 2 &  [0,1] (1)\\
\bottomrule
\end{tabularx}
\begin{tablenotes}
    \item  G= group fairness and I = individual fairness. P= Actual Positives, N= Actual Negatives, TN= True Negatives, TP= True Positives, FP= False Positives, FN= False Negatives. The WGEI and WGTI are derived from the generalized entropy index \citep{spei} given by $\varepsilon^\alpha (b_1,b_2,....,b_n)=\frac{1}{n\alpha(\alpha - 1)}\sum_{i=1}^{n}[(\frac{b_i}{\mu})^{\alpha}-1] = \varepsilon^{\alpha}_{\beta}(\textbf{b}) + \varepsilon^{\alpha}_{\omega}(\textbf{b})$. Where the \textit{benefit} for individual $i$ is derived from a function that assigns a score signifying how much an algorithmic decision benefits an individual via $b_i= \hat{y}_i - y_i +1$. $\alpha$ is a constant and $\varepsilon(.)$ is the entropy of benefits. 
\end{tablenotes}
\end{threeparttable}
\end{table*}
\subsection{Fairness Algorithms}
Fairness algorithms are often designed to facilitate ML models such that the ML models satisfy one or more fairness metrics. In such a case, the ML model is said to be ``fair''. Seven well known fairness algorithms covering pre-processing, in-processing and post-processing within an ML pipeline have been considered in this work.

\subsubsection{Pre-processing Algorithms}
Pre-processing algorithms tackle algorithmic unfairness by debiasing data. The fair data generated by the pre-processing algorithm can then be used to train any downstream ML model. The 3 pre-processing techniques we used in this study are Suppression (SUP) \citep{kamiran12}, Reweighing (RW) \citep{kamiran12}, and Disparate Impact Remover (DIR) \citep{feldman2015certifying}. 
\subsubsection{In-processing Algorithms}
In-processing algorithms achieve fairness by explicitly introducing fairness constraints in the ML algorithm. We consider 2 in-processing algorithms in this study namely Prejudice Remover (PR) \citep{kamishima12} and Advesarial Debiasing (AdDeb) \citep{zhang18}. 
\subsubsection{Post-Processing Algorithms} Post-processing methods involve altering the outcomes of a pre-trained model to attain specific fairness criteria across various groups. The 2 post-processing algorithms considered in this study are Equal Odds algorithm (EQ) \citep{hardt16} and Calibrated Equal Odds (CEq) \citep{pleiss}.

\section{Experimental Setup}
\label{sec:experiment}
\subsection{Datasets and Baseline ML Algorithms}
We used 5 real-world datasets for our comparative analysis. Three of the datasets are commonly used publicly available datasets\footnote{\href{https://t.ly/1-Gn2}{https://t.ly/1-Gn2}}\footnote{\href{https://shorturl.at/wBDIR}{BAF data available here}} in the algorithmic fairness community while two are proprietary datasets. The two proprietary datasets are anonymized \textit{counts} of first semester Moodle\footnote{Moodle is a virtual learning platform} engagement records for 2 mandatory STEM courses---coded as NWF and ITF---in a large public Australian university. Table \ref{tab:datasets} presents a summary of all the datasets used, which are also briefly described as follows. \textbf{Proprietary Datasets}: NWF and ITF represent students' demographic and engagement records from 2015 to 2020 for two courses respectively. \textbf{Public Datasets}: BAF is synthetic data generated from anonymized real-world bank fraud detection dataset spanning 8 months \citep{jesus2022turning}. The Adult dataset is a income census dataset detailing whether a person's income exceeds 50K US dollars \citep{misc_adult_2}. The COMPAS dataset contains criminal history and COMPAS risk scores for defendants from Broward County \citep{angwin16}. 
\begin{table*}[!htbp]
\caption{Description of all datasets. The seasons show how a dataset was partitioned into the 3 seasons format}
\label{tab:datasets}
\centering
\tiny
\begin{threeparttable}
\begin{tabularx}{\textwidth}{lllllllllll}
\toprule
Data & Size & Features & Domain & PA & TS  & Public & Outcome &\multicolumn{3}{c}{Season} \\
\cmidrule(l){9-11}
 &  &   &  &  &   &  & & $t_0$ & $t_1$ & $t_2$ \\
\midrule
NWF& 1,427 &  11 &  Edu& Ctzn& \checkmark &  N/A&  P/F &  2015/16 & 2017/19  & 2020 \\
ITF& 1,246 &  11 &  Edu& Ctzn& \checkmark &  N/A&  P/F & 2015/16 & 2017/19  & 2020\\
BAF & 319,672 &  24 &  Fin& Age& \checkmark &  \checkmark&  F/NF & Month0-2 & Month6  & Month7\\
Adult & 45,063 &  14 &  Emp & Gender & N/A &  \checkmark &  I> 50K & Original & Drift1  & Drift2\\
CMP & 6,172 &  13 &  CJ & Race & N/A &  \checkmark &  2y Rcd & Original & Drift1  & Drift2\\
\bottomrule
\end{tabularx}
\begin{tablenotes}
    \item CMP= COMPAS dataset. Ctzn = Citizenship. Edu = Education, Fin = Finance, Emp = Employment, and CJ= Criminal Justice. PA= Protected Attribute, TS= Timestamp. P/F = Pass/Fail, F/NF= Fraud/No Fraud, I >50K = Income > 50K, and 2y Rcd= two-year recidivism. 
\end{tablenotes}
\end{threeparttable}
\end{table*}

For the baseline algorithms, i.e., the algorithms without any fairness constraint, we consider 4 classical ML algorithms that have been extensively used in several related works as baselines for comparing fairness algorithms, namely Logistic Regression (LR), eXtreme Gradient Boosted Trees (XGB), Random Forest (RF), and Support Vector Machines (SVM) \citep{islam2022through}.

\subsection{Analysis of Covariate Drift in Datasets}
Given that the important covariates have significant influence on the prediction outcomes \citep{zien2009feature}, we are interested in investigating the relationship between covariate importance, covariate drift, and algorithmic unfairness. To that end, we pursued two key objectives. Firstly, we are interested in knowing if a significant drift in the important covariates will correspond to significant levels of unfairness. Secondly, we are interested in knowing if unfairness flows in the direction of the drift of the important covariates. Specifically, we want to investigate whether, if the most important covariates drifts significantly for a particular demographic group, the measured bias will also flow in that same direction.
\par \textbf{Ranking of Covariates.} To determine the important covariates, for each dataset, we trained the 4 baseline models, i.e., RF, LR, XGB, and SVM and used the co-efficient weights (for LR and SVM), covariate importance scores (for XGB and RF), and the well-established SHapley Additive exPlanations (SHAP) values (for all 4 models). All the covariate importance scores were normalized to be between 0 and 1 and used to rank the covariates. Ranking results have been excluded due to space constraints. The supplementary results of \textit{all} experiments in this paper are available at \href{https://shorturl.at/4ErID}{\textcolor{blue}{ this link}}.

\par \textbf{Covariate Drift: } All datasets were partitioned into 3 seasons based on the timestamp of the records in each dataset. Thus simulating historical training dataset at time $t_0$ denoted, and two test datasets at times $t_1$ and $t_2$ respectively. For example, the NWF dataset is partitioned as (1) Historical data at time $t_0$ (2015-2017), i.e., \textit{l}ong \textit{b}efore \textit{C}OVID-19 (LBC); (2) Pre-covid data at time $t_1$ (2018-2019), i.e., immediate \textit{p}re-\textit{C}OVID-19 (PC); and (3) Peri-covid data at time $t_2$ (2020), i.e., during or \textit{pe}ri-\textit{C}OVID (PeC). We computed the size of covariate drift across the 3 seasons using Jensen-Shannon Distance (JSD) in the following fashion: $JSD_{t_{01}}(X_{t_0},X_{t_1})$ and $JSD_{t_{02}}(X_{t_0},X_{t_2})$  (c.f. Equation \ref{eq:jsd}). Figure \ref{fig:NWFdrift} and Figure \ref{fig:BAFdrift} show the covariate drift patterns of the NWF and BAF datasets. It can be observed that the figures show an inconsistent mix of gradual and sudden drifts. This observation is important as it demonstrates that there can be different levels of drift, which require tailored attention, unlike the works of \citep{islam2022through, ghodsi2022context} which assume the same level of drift and thus treat the impact of drift to fairness equally. We noted that the Adult and COMPAS datasets did not have timestamped records. In this case we first set the \textit{original} dataset as the \textit{Historical} dataset at season $t_0$.  We then introduced a calculated \textit{artificial drift} to the most important covariates and denote this dataset as \textit{Pre-covid} equivalent at season $t_1$. We did the same for the \textit{Peri-covid} equivalent at season $t_2$. We introduced the calculated drifts randomly across demographic groups. We refer to these two drifted datasets as \textit{Drift 1} and \textit{Drift 2} respectively.  For instance, Figure \ref{fig: Compasdrift} shows the introduced drift for the COMPAS dataset.  We created the artificial drift using the following equation:
\begin{equation}
\label{eqn:drift}
    \text{Artificial Drift} = \bar X \cdot k + c
\end{equation}
\begin{where}
\item  $\Bar{X}$ is the mean of the covariate; $c \in [0.001,0.1]$ is small constant to ensure non-zero drift; and $k \in [0,3]$ is the level of drift to be introduced.
\end{where}
\subsection{Model Training and Testing} We refer to the models without any fairness constraints as \textit{baseline models}, and the models with fairness constraints as \textit{fairness-aware models}. Overall, we used a total 4 baseline models and 22\footnote{We obtained the 22 fairness-aware models from the various combinations of pre, and post-processing algorithms with the baseline models} fairness-aware models for all our experiments as shown in Table \ref{tab:all_models}.\\
\textbf{Train-Test Split}: Recall that all datasets are partitioned into 3 seasons at times $t_0$, $t_1$, and $t_2$ which represent the \textit{historical} training dataset, the test dataset$_1$, and the test dataset$_2$ respectively. To compare the impact of the implicit \textit{iid} assumption often made in the literature, we used cross-validation 
 and bootstrapping to simulate temporal impact of the \textit{iid} assumption as follows. In the cross-validation, we performed 40 shuffled repeats of 5-fold cross validation, making a total of 200 train-test runs on the historical training dataset. The results from this cross-validation represents situation where the training and test datasets are drawn from the same distribution, i.e., the \textit{iid assumption}. In the bootstrapping, we trained each model on 200 bootstrapped samples of the historical dataset and tested them on  the test dataset$_1$ and test dataset$_2$. The results from this bootstrapping experiment represent the scenario where the training and test datasets do not follow the \textit{iid} assumption. All hyper-parameters where optimized accordingly.

\section{Results and Discussion}
\label{sec:resdisc}
\begin{figure}
    \centering
        \includegraphics[width=\linewidth]{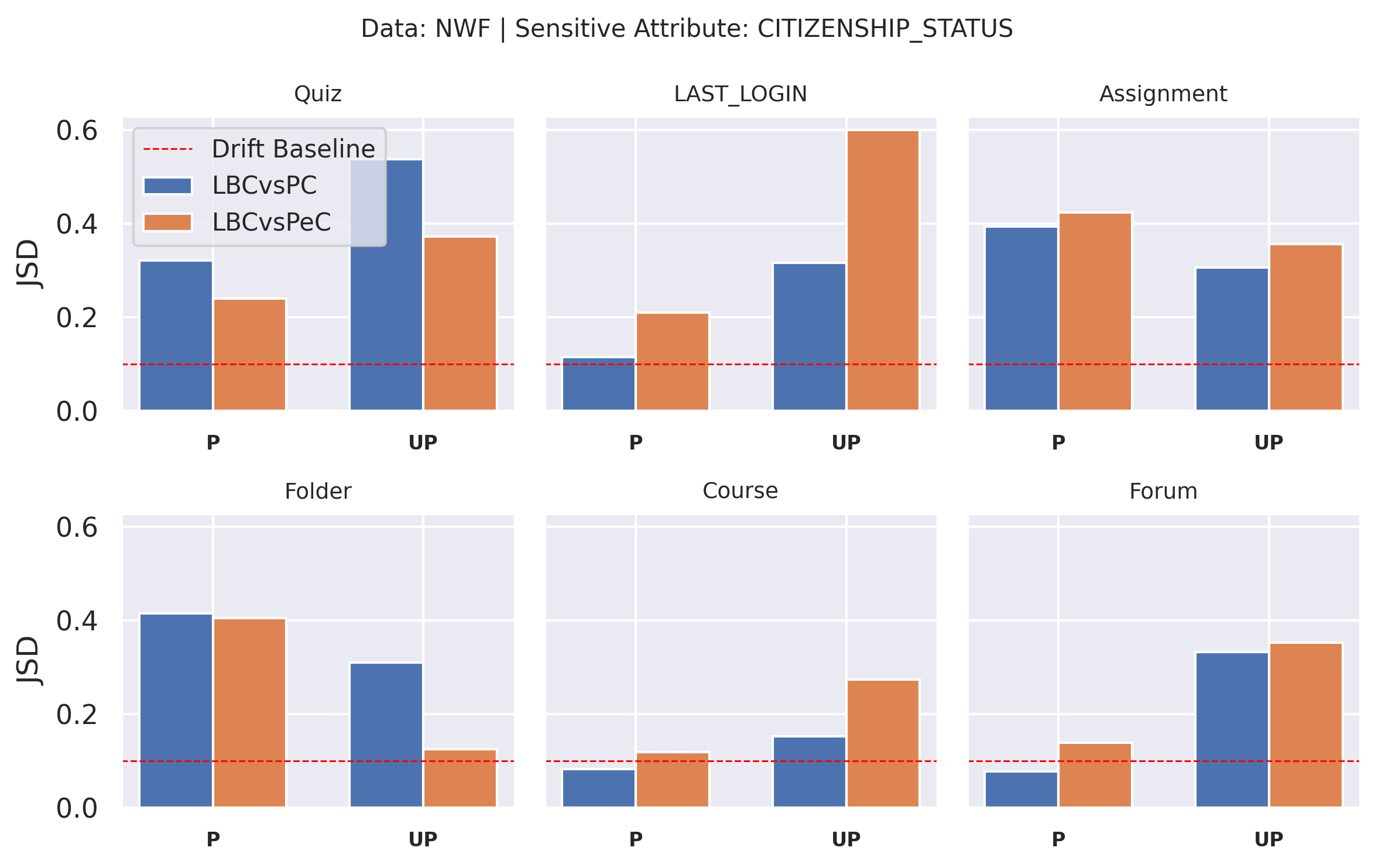}
    \caption{Drifts of top-6 important covariates for NWF dataset. P = privileged group, UP = unprivileged group . LBC= long before covid ($t_0$). PC = pre-covid ($t_1$), and PeC = peri-covid ($t_2$).}
    \label{fig:NWFdrift}
\end{figure}

\begin{figure}
    \centering
      \includegraphics[width=\linewidth]{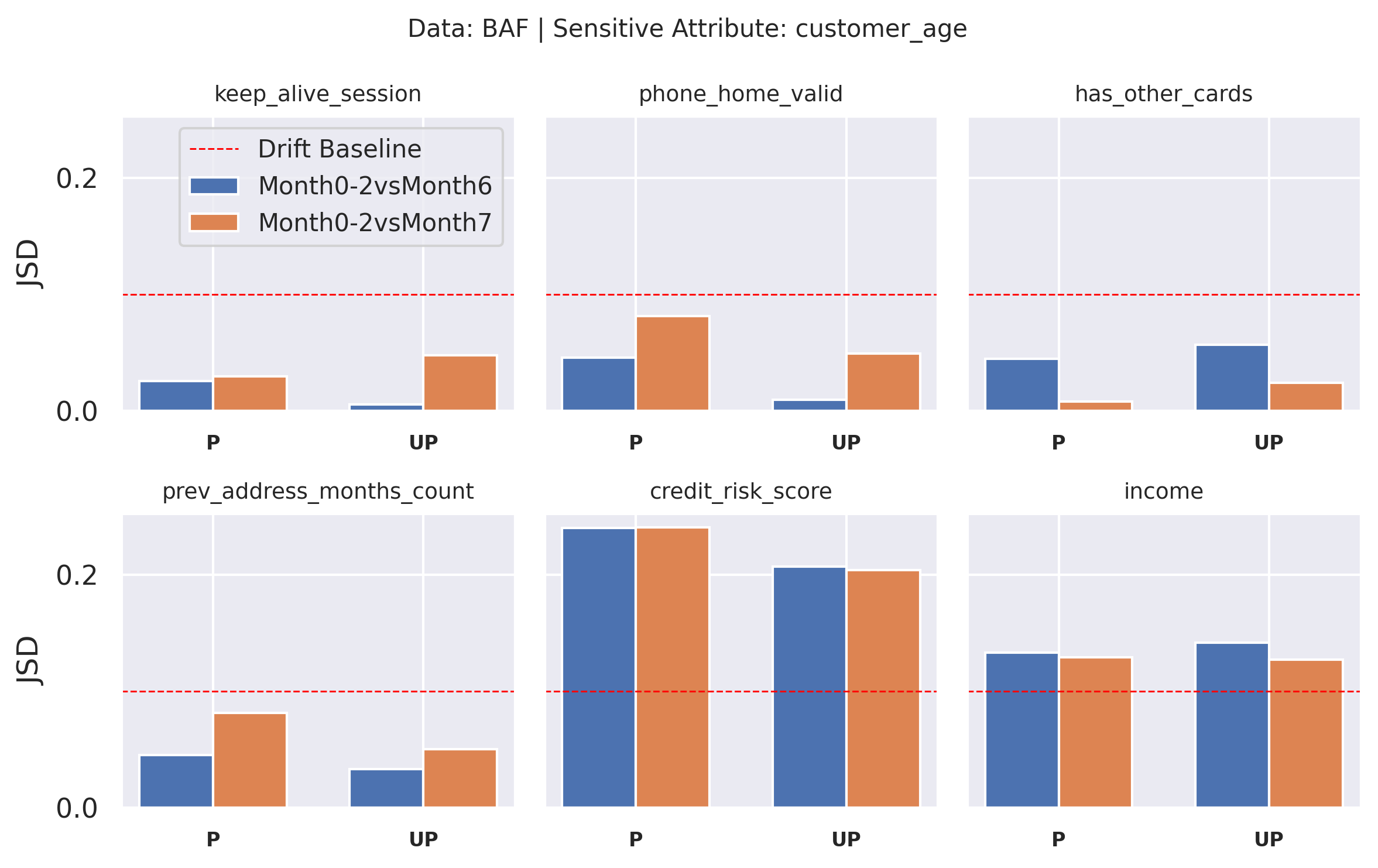}
    \caption{Drifts of top-6 important covariates for BAF dataset.}
    \label{fig:BAFdrift}
\end{figure}

 \begin{figure}
    \centering
      \includegraphics[width=\linewidth]{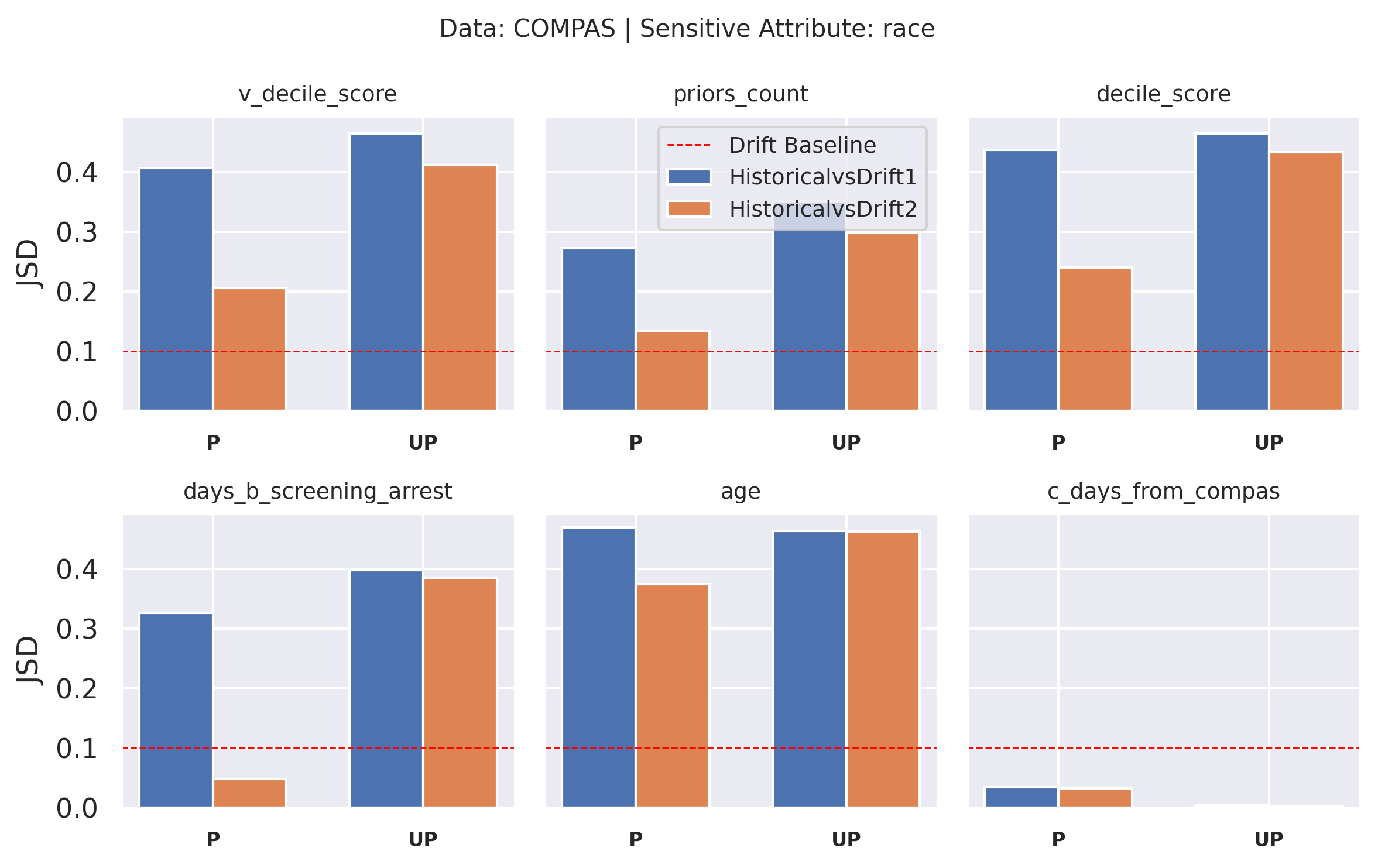}
    \caption{Drifts of top-6 important covariates for COMPAS dataset.}
    \label{fig: Compasdrift}
\end{figure}

\begin{table}
\caption{Configurations of fairness-aware models w.r.t to baseline models}
\centering
\footnotesize
\begin{tabularx}{.6\linewidth}{cc}
\toprule
\textbf{Type} & \textbf{Model} \\
\midrule
\multirow{1}{*}{Baseline} &  (XGB, RF, LR, SVM) \\
\hline
\multirow{3}{*}{Pre-processing} & SUP+ (XGB, RF, LR, SVM) \\
& RW+ (XGB, RF, LR, SVM) \\
& DIR+ (XGB, RF, LR, SVM) \\
\hline
\multirow{1}{*}{In-Processing} & PR, AdDeb \\
\hline
\multirow{2}{*}{Post-processing} & EQ+ (XGB, RF, LR, SVM)\\
& CEq+ (XGB, RF, LR, SVM)\\
\bottomrule
\end{tabularx}
\label{tab:all_models}
\end{table}

\begin{figure}
    \centering
        \includegraphics[width=\linewidth]{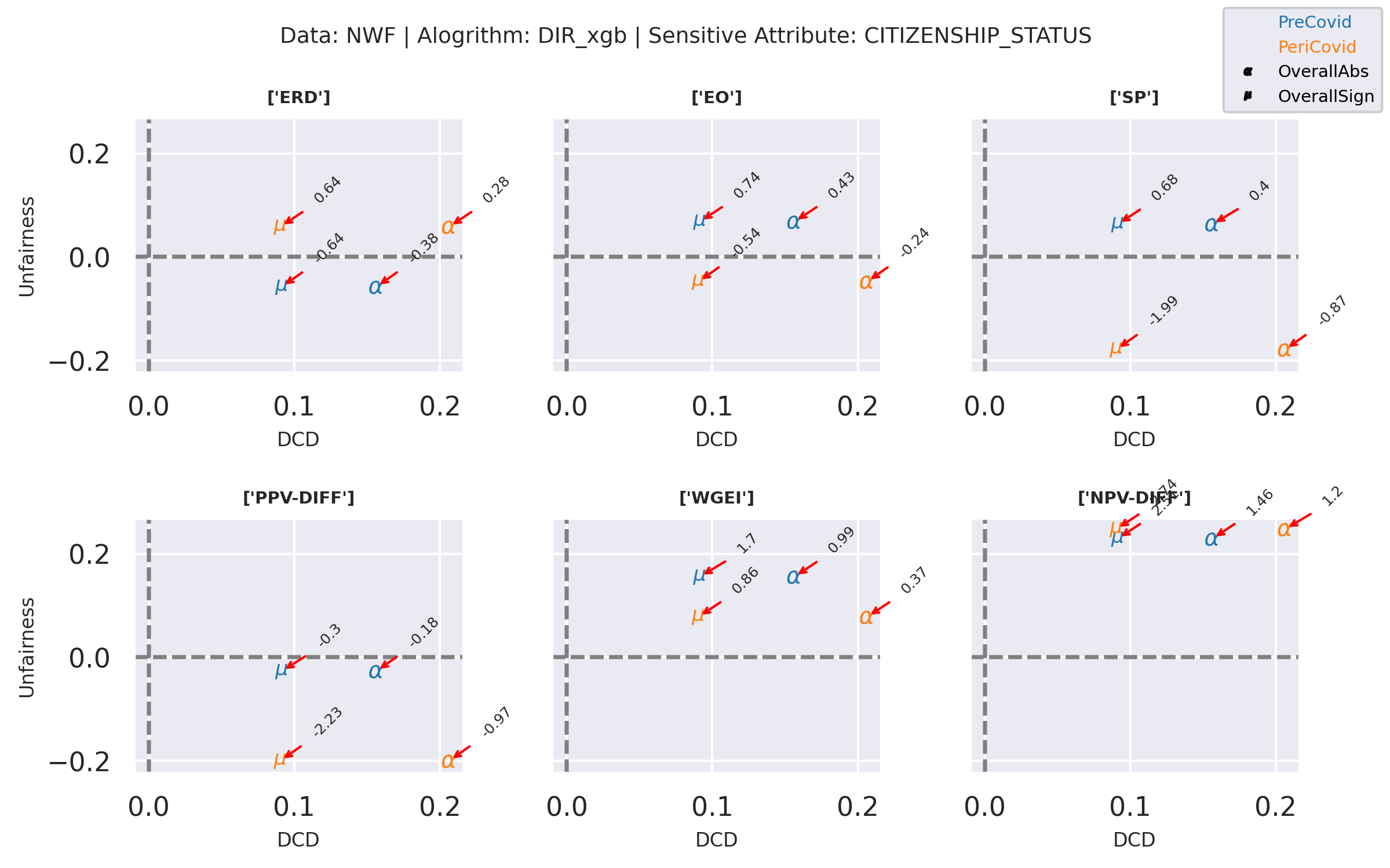}
    \caption{DCD vs unfairness for DIR+XGB model for NWF dataset. Results are similar for all models, metrics and datasets.}
    \label{fig:NWFdriftvsfairness}
\end{figure}

\begin{figure}
    \centering
      \includegraphics[width=\linewidth]{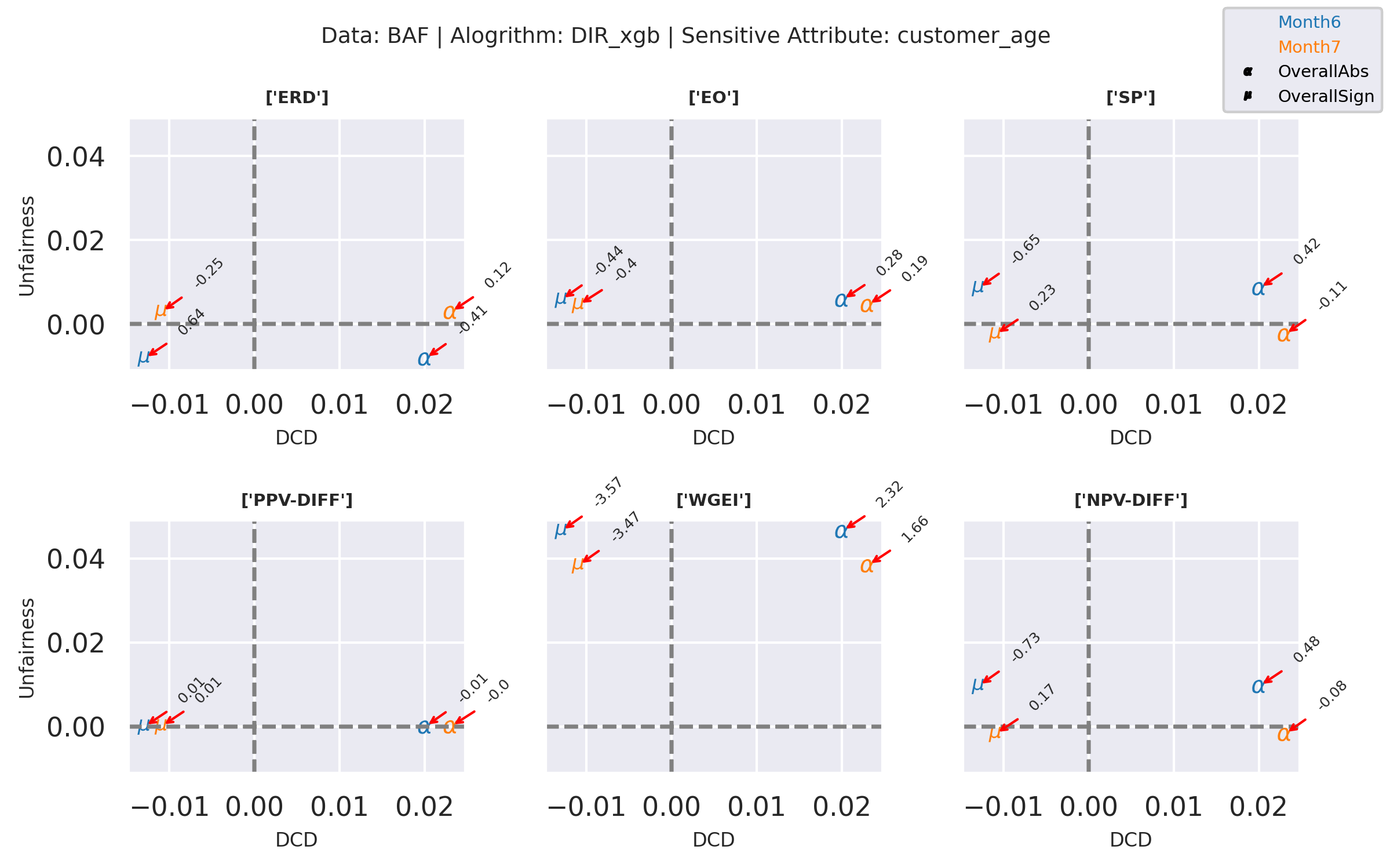}
    \caption{DCD vs unfairness for DIR+XGB model for BAF dataset.}
    \label{fig:BAFdriftvsfairness}
\end{figure}
\subsection{Relationship Between Differential Covariate Drift and Fairness}
Differential Covariate Drift (DCD), which is the difference between the covariate drift of a privileged group and an unprivileged group, is first calculated. We then computed the overall DCD by finding the \textit{signed} and \textit{absolute} average of the DCDs in the top-6 most important covariates\footnote{We chose the top-6 important covariates since they had the most predictive utility when we did the normalized covariate ranking for majority of the datasets}. From hereon, we simply refer to the overall DCD as DCD. To establish the relationship between DCD and fairness, we computed the rate of the change of DCD wrt fairness. Figure \ref{fig:NWFdriftvsfairness} and Figure \ref{fig:BAFdriftvsfairness} show the relationship between DCD and fairness on the NWF and BAF datasets respectively. In both figures, $\alpha$ and $\mu$ represent the absolute and signed DCDs respectively, and the annotations represent the gradient from the origin (0,0). Furthermore, in the x-y plane, all the negative coordinates are against the unprivileged group and the positive coordinates are against the privileged group. In both figures, it can be seen that the relationship between DCD and fairness is not straightforward. 
\par Firstly, we observed that \textit{unfairness does not always flow in the direction of the DCD}. In other words, if the unprivileged group, for example, has a higher DCD, it does not \textit{necessarily} mean that the models will be biased in favour of the unprivileged group. For example, consider Figure \ref{fig:NWFdriftvsfairness}. In terms of the ERD metric, we observed that the model was biased \textit{for} the unprivileged (ERD = 0.07) Peri-covid. Consistently, we observed that the absolute DCD for Peri-covid (i.e., $t_0$ vs. $t_2$) was also higher for the unprivileged (DCD = 0.2). In contrast, for the same ERD metric, Pre-covid, we observed that the model was biased \textit{against} the unprivileged group (ERD = -0.06) even when the absolute DCD for Pre-covid (i.e., $t_0$ vs. $t_1$) was still higher for the the unprivileged groups (DCD = 0.15). We made similar observations across all metrics and all models---both baseline and fairness-aware. \textbf{Key takeaway: }\textit{Across all datasets, fairness metrics and models, we show that positive (resp. negative) DCD does not imply positive (resp. negative) discrimination}. 
\par Secondly, we observed that \textit{the size of the DCD is not always proportional to size of the unfairness}. That is, given two DCDs, the lower DCD does not necessarily imply lower unfairness and vice versa. For instance, in Figure \ref{fig:NWFdriftvsfairness}, for the EO metric, the absolute DCD for Peri-covid is higher (0.2) than the DCD for Pre-covid (0.15), however, we observed that the unfairness in the EO metric for Peri-covid is lower (0.05) than that of the Pre-covid (0.07). Similar observations were made for different models across different datasets (e.g., see Figure \ref{fig:BAFdriftvsfairness}). \textbf{Key takeaway:} \textit{Differential covariate drift may not always cause unfairness. This finding is consistent with \cite{gardner2023cross} and contrary to that of \cite{castelnovo2021towards}}

\par Furthermore, we observed that the \textit{significance of the drift in the important covariates is indicative of its consequent impact on algorithmic unfairness}. Consider Figure \ref{fig:NWFdrift} and Figure \ref{fig:BAFdrift} which represent covariate drifts across privileged and unprivileged demographic groups for the NWF and BAF datasets respectively. In these figures, the red-dashed lines represent the \textit{baseline drift} above which covariate drifts are considered significant. The blue bars represent the covariate drifts at times $t_0$ vs. $t_1$, and the orange bars represent covariate drifts at times $t_0$ vs. $t_2$. The top-6 covariates are shown in order of importance row-wise, from left to right.  For the NWF, that is Figure \ref{fig:NWFdrift}, it can be clearly seen that the most important covariates, e.g., last login and quiz, significantly drifted. On top of that, the DCDs were relatively high---especially for the unprivileged group. Consequently, unfairness was high across all fairness metrics as shown in Figure \ref{fig:NWFdriftvsfairness}. Indeed, unfairness could be as high as 0.28 (28\%) in terms of the NPV-DIFF metric. In contrast, consider the Figure \ref{fig:BAFdrift} for the BAF dataset. We observed that the top-4 most  important covariates showed \textit{insignificant} drifts. In addition to the insignificance of the drifts, the DCDs were very little. In fact, the highest DCD for the BAF dataset was approximately a measly -0.03 (3\%) as compared the DCD of NWF which could be as high as 0.2 (20\%). Consequently, we observed that the models trained on the BAF dataset displayed little unfairness. Similar observations were made across the other datasets. \textbf{Key takeaway:} \textit{Significant drifts coupled with high DCD in important covariates is likely to result in higher algorithmic unfairness.}

\subsection{Robustness of Fairness Algorithms to Covariate Drift}
This section focuses on the ability of fairness-aware models to maintain their predictive accuracy and fairness guarantees in the presence of covariate drift. Figure \ref{fig:fairBAF_EO} and  Figure \ref{fig:fairITF_SP} are the results of this experiment in terms of equalized odds statistical parity for the BAF and ITF datasets respectively. In both figures, the green-dashed line indicates the ideal fairness score. The gray bars indicate the fairness score when the training and test datasets are from the same distribution at $t_0$, thus following the \textit{iid} assumption. The purple and olive bars indicate the fairness scores for training dataset at $t_0$ and test datasets at $t_1$ and $t_2$ respectively. The purple and olive bars represents the non-\textit{iid} assumption. Negative fairness scores represent unfairness against the unprivileged group and positive scores indicate otherwise. Furthermore, the annotations on the bars are the average weighted F1-scores for each model on each test dataset. The bar heights are the average of the particular fairness metric and the error bars are the standard errors of the means. We made the following observations.
\begin{figure}
    \centering
      \includegraphics[width=\columnwidth]{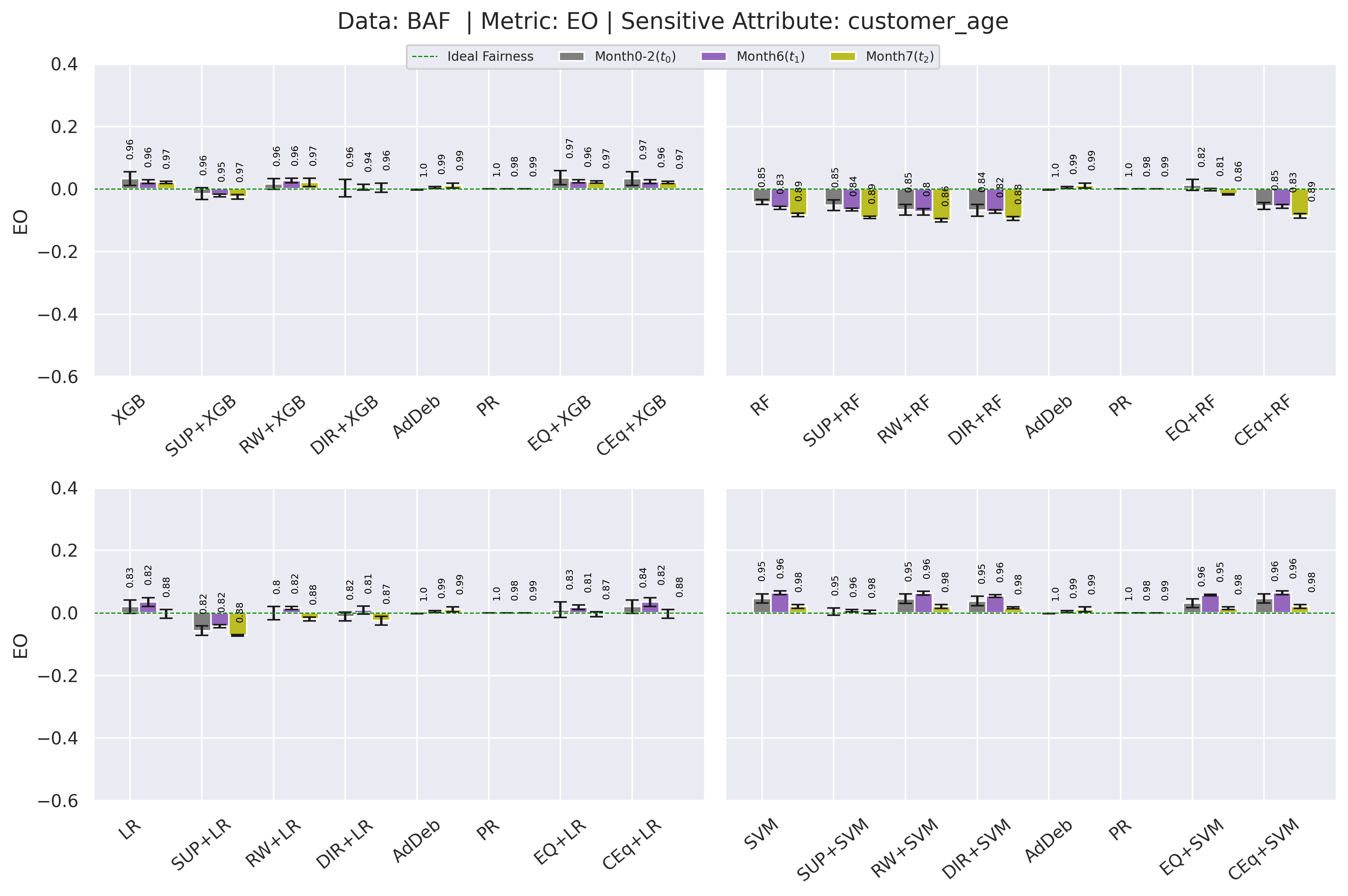}
    \caption{Robustness of baseline and fairness-aware models on the BAF dataset in terms of EO metric.}
    \label{fig:fairBAF_EO}
\end{figure}
\par In both Figure \ref{fig:fairBAF_EO} and  Figure \ref{fig:fairITF_SP}, we observed that \textit{none of the fairness-aware models that we evaluated is consistently robust}. None of the fairness-aware models we investigated was consistently able to maintain their fairness and predictive capability in the face of covariate drift across different metrics and datasets. Nevertheless, we observed that certain algorithms performed better than other algorithms on certain datasets at times. For instance, on the BAF dataset, we observed that the PR algorithm consistently achieved near perfect fairness in the EO metric with minimal drop in predictive accuracy as shown in Figure \ref{fig:fairBAF_EO}. In fact, except for the \textit{consistency} metric which we found to be insensitive to the fairness-aware models just like \cite{majumder2021fair} did, we found that the PR algorithm consistently performed well in terms of all predictive accuracy and fairness metrics at times $t_0$ through to $t_2$. It is worth noting that the BAF dataset is highly imbalanced in the target class and the demographic groups have equal base rates. In general, we observed that the in-processing approaches, specifically the PR algorithm, tend to be relatively robust as compared to the pre- and post-processing approaches which partly contradicts the findings of \cite{islam2022through}. We made similar observations across other datasets and fairness metrics, except in terms of the \textit{consistency} metric which was insensitive. \textbf{Key takeaway:} \textit{Existing fairness-aware models are often not able maintain their utility in the presence of covariate drift across different fairness metrics and datasets}
\par We also observed that \textit{the fairness of an algorithm is a function of time}. We found that claims of fairness of an algorithm on a static snapshot of data---as is commonly done in literarture---can be misleading. We made this observations across datasets for different fairness metrics. For instance, consider the Figure \ref{fig:fairITF_SP} for the ITF dataset. In the fourth quadrant (clockwise), it can be clearly seen that the DIR+XGB was the least fair model in terms of the SP metric at $t_0$, and even at $t_1$. However, at $t_2$, the DIR+XGB was the most fairest. Similar trends can be observed in the other quadrants. Indeed our results suggest that a \textit{supposedly} unfair algorithm, might actually be fair, using the same metric, in the same domain, but on (test) dataset from a different time. The reverse of this logic also holds. Therefore, we posit that concluding a fairness approach as fair or unfair should be contextualized. \textbf{Key takeaway:} \textit{Unless claims of fairness or unfairness are contextualized e.g., wrt time, conclusions about a particular fairness algorithm's ``superiority'' can be misleading.}
\begin{figure}
    \centering
      \includegraphics[width=\columnwidth]{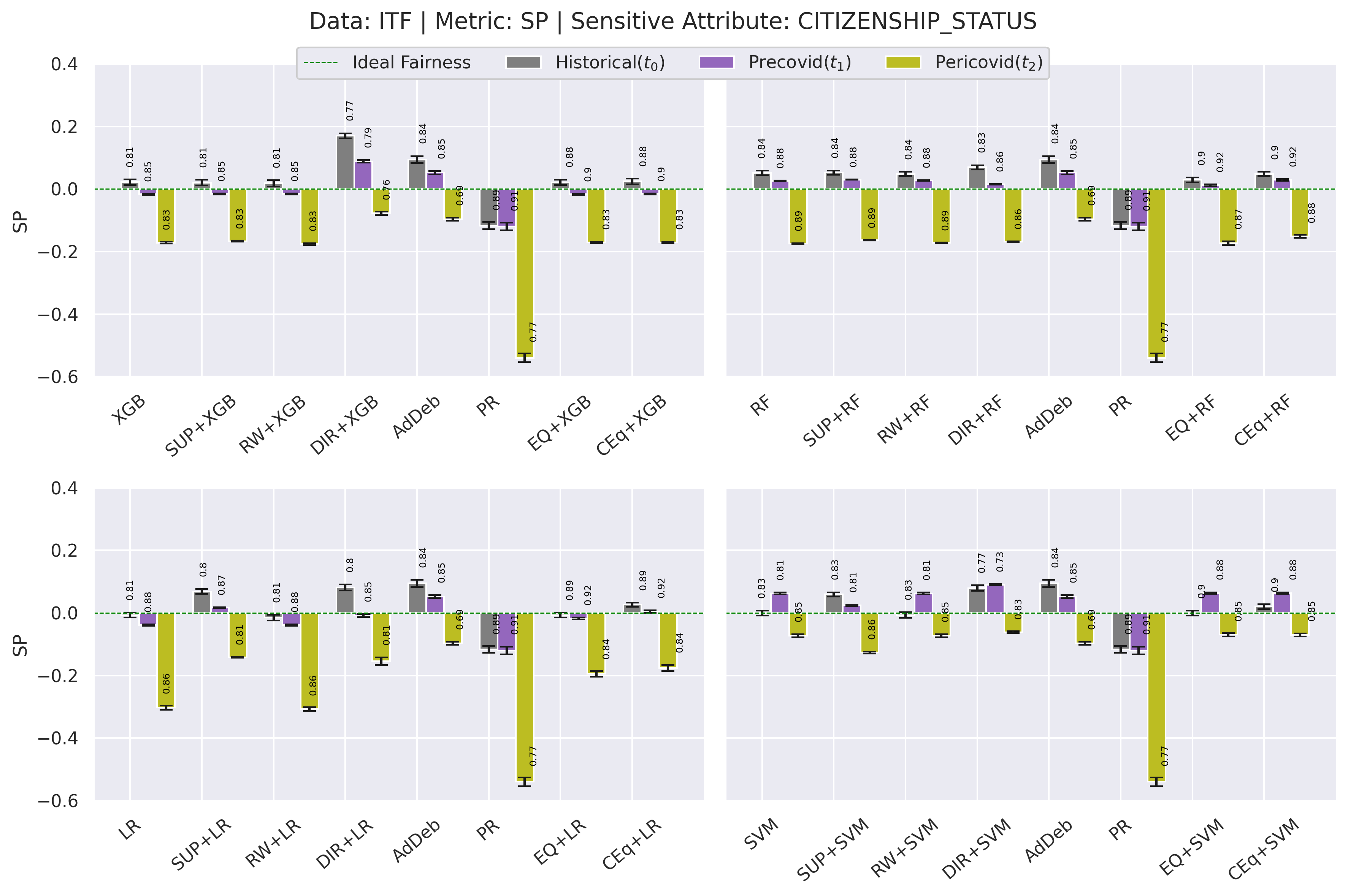}
    \caption{Robustness of baseline and fairness-aware models on the ITF dataset in terms of SP metric.}
    \label{fig:fairITF_SP}
\end{figure}
\subsection{Impact of Design Choices on Fairness}
We made other interesting observations based on design choices such as hyper-parameter optimization and baseline model selection across all datasets and fairness metrics. For example, consider Figure \ref{fig:fairBAF_EO} which represents the fairness of both baseline and fairness-aware models in terms of the EO metric on the BAF dataset at times $t_0$ to $t_2$. Given the same pre-processed data from the DIR algorithm, DIR+XGB seems to be the fairest whereas the DIR+RF seems to be the least fair at times $t_0$, $t_1$, and $t_2$ . In fact, the DIR+RF was discriminatory \textit{against} the unprivileged group whereas the DIR+SVM appeared to discriminate \textit{for} the unprivileged. We made similar observations for the post-processing approaches that were run on top of the ML models. Furthermore, whereas \cite{islam2022through} observed that the post-processing approaches tend to be generally less be impacted by the choice of ML model as compared to the pre-processing, we observed that neither approach consistently has an edge over the other. Similar to \cite{roth2018comparison}, we rather found the in-processing approaches, particularly the PR algorithm to be less impacted since they are tied to a single model at design. \textbf{Key takeaway:} \textit{The choice of downstream ML model is a key indicator of the fairness and predictive accuracy for pre- and post-processing algorithms, thus should be carefully considered.}
\subsection{Implications for practice}
From our findings, we discuss some useful practical implications. 
\par \textit{There is the need for continuous monitoring and evaluation of fairness algorithms}. An implication of this is that, before classifying a fairness algorithm as fair and fit for deployment, or classifying an algorithm as unfair and should be discarded, fairness practitioners may have to continuously monitor the deployed algorithm as the \textit{tides of fairness can turn really fast}. A few recent works have started to explore this line of research \citep{ henzinger2023runtime,henzinger2023monitoring}. This is an emerging research area that requires significant attention. 
\par \textit{The flexibility (e.g., in choice of base ML models) in pre- and post-processing approaches can be both strength and weakness}. The ability of pre- and post processing approaches to pair up with any downstream models gives them the convenience of variety. Moreover, pre- and post-post processing approaches allow seamless integration with popular and powerful ML libraries such as sci-kit learn. However, the downstream models that are applied to the fair data are not designed with any fairness constraints. Therefore, the fairness of  the downstream models are not guaranteed. Moreover, extra customization and hyper-parameter optimizations may undo whatever fairness was incorporated in the fair data. Additionally, pre-processing approaches mostly aim to correct the ground truth labels. Therefore, predictive error or accuracy-based fairness metrics cannot be catered for by pre-processing approaches \citep{islam2022through}. Fairness practitioners may address this weakness by ensembling pre- and in-processing approaches where the flexiblity of pre-processing is hybridized with the strictness of in-processing approaches. 
\par \textit{The cause of algorithmic unfairness is a cocktail of latent variables}. Differential covariate drift, is without a doubt, a source of algorithmic unfairness, thus should be addressed. However, wrongly attributing unfairness to differential covariate drift may cause the unfairness to persist even if the algorithm is monitored round-the-clock for any drift and appropriately handled. The source of algorithmic unfairness is multifaceted. Therefore, fairness researchers may have to identify the source of unfairness on a case by case basis and address it accordingly instead of proffering a one-size-fits-all  solution.  
 
\section{Conclusion}
\label{sec:conc}
In this work, we investigated the relationship between differential covariate drift and algorithmic unfairness, and we further analyzed the robustness of 7 existing fairness algorithms in the face of covariate drift. We found that significant drifts in important covariates in addition to higher differential covariate drifts often leads to unfairness. We also found that none of the existing fairness-aware algorithms that we evaluated are robust in the presence of covariate drift. Even more interestingly, in contrast to certain prior studies, we found that there is no correlation between the magnitude and direction of data distributional drift and the ensuing level and direction of unfairness.  Based on these insights, the study offers policy implications related to the impact of data distributional drift on fairness algorithms. These implications are important for relevant stakeholders, offering valuable guidance on addressing and mitigating fairness issues in the presence of covariate drift.
\par Recently, there are some algorithms that have been designed with distributional drift in mind \citep{chen2022fairness,du2021fair,taskesen2020distributionally,robust}. In a future study, we intend to perform similar investigations on these algorithms to ascertain if they are \textit{indeed} robust to distributional drift as claimed. Furthermore, we intend to investigate an interesting line of research which suggests that stability of fairness algorithms can be achieved via surrogate functions by reducing surrogate fairness-gaps and variance \citep{yao2024understanding}.

\backmatter

\bmhead{Supplementary information}
Supplementary results are available online at \href{https://shorturl.at/4ErID}{\textcolor{blue}{ this link}} 

\bmhead{Acknowledgements}

This work has been supported by the Food Agility Cooperative Research Center, Australia.

\section*{Declarations}
\textbf{Conflict of interest}: The authors have no Conflict of interest to declare that are relevant to the content of this article.\\
\textbf{Ethics approval}: The study was approved by the governing ethical board (Ethics Protocol Application ID: 204198). 

\bibliography{bibliography}

\end{document}